\documentclass{article}
\usepackage{spconf,amsmath,graphicx}
\usepackage{subfig}
\usepackage{multirow}
\usepackage{hhline}
\usepackage{pgf}
\usepackage{tikz}
\usepackage{tikzscale}
\usepackage{url}
\usetikzlibrary{arrows,automata, positioning, fit, backgrounds}

\def\hptr{{HPTR}}

\def\cihpc{{CCIHP}}
\def\hp{{HP}}
\def\appv{$AP^p_{vol}$}
\def\aprv{$AP^r_{vol}$}
\def\apcrv{$AP^{cr}_{vol}$}

\usepackage{caption}

\usepackage{fancyhdr}

\title{Describe Me If You Can! 
\\ Characterized Instance-level Human Parsing}
\name{Angelique Loesch$^{\star, \dagger}$ \qquad Romaric Audigier$^{\star, \dagger}$
}

\address{$^{\star}$ Université Paris-Saclay, CEA, List, F-91120, Palaiseau, France\\
    $^{\dagger}$Vision Lab, ThereSIS, Thales SIX GTS, Campus Polytechnique, Palaiseau, France\\
    \{angelique.loesch, romaric.audigier\}@cea.fr\\}
\begin{document}
%
\maketitle
\begin{abstract}
Several computer vision applications such as person search or online fashion rely on human description. 
The use of instance-level human parsing (\hp{}) is therefore relevant since 
it localizes
semantic attributes and body parts within a person.
But how to characterize these attributes?
To our knowledge, only some single-\hp{} datasets 
describe attributes with some color, size and/or pattern characteristics. 
There is a lack of dataset for multi-\hp{} in the wild with such characteristics. 
In this article, we propose the dataset \cihpc{} based on the multi-\hp{} dataset CIHP, with 20 new labels covering these 3 kinds of characteristics.\footnote{\cihpc{} is available on https://kalisteo.cea.fr/index.php/free-resources/}
In addition, we propose \hptr{}, a new bottom-up multi-task method based on transformers as a fast and scalable baseline.
It is the fastest method of multi-\hp{} state of the art while having precision comparable to the most precise bottom-up method.
We hope this will encourage research for fast and accurate methods of precise human descriptions.

\end{abstract}
\vspace{-0.1cm}
\section{Introduction}
\label{sec:intro}

Human semantic description is of utmost importance in many computer vision applications.
It consists in automatically extracting semantic attributes corresponding to each person of an image.
Attribute extraction is useful for many types of tasks, as \emph{image content description}, \emph{image generation} for virtual reality applications or \emph{person retrieval} from a natural-description query, for security applications.
Semantic attributes can also
help visual signatures used in 
person \emph{re-identification} and \emph{person search} pipelines \cite{Loesch2019,Lin2019}. 

Unlike \emph{attribute classification} that aims to predict multiple tags to the image of a person~\cite{Lin2019} (sometimes deceived by nearby people/elements),
\emph{human parsing (\hp{})}~\cite{Gong2019, Ruan2019, Yang2020b, Gong2018} aims to segment visible body parts, clothing and accessories at the pixel level.
Localizing semantic attributes has several advantages.  
It provides a precise delineation of attributes necessary in augmented/virtual reality applications (entertainment, clothing retail...) \cite{Hsieh2019}.
It is more explainable than global tags (thus, more acceptable by human operators) and can cope with multiple-person descriptions by directly assigning localized people with attributes.

%

\begin{figure}[!t]
  \centering
  \includegraphics[width=0.25\linewidth]{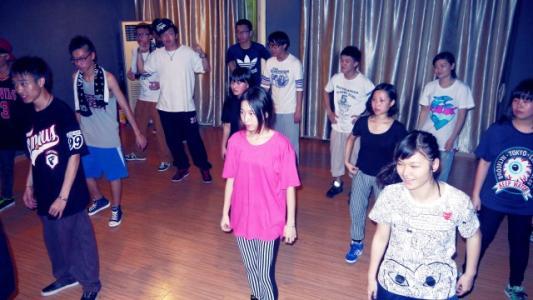}
  \includegraphics[width=0.25\linewidth]{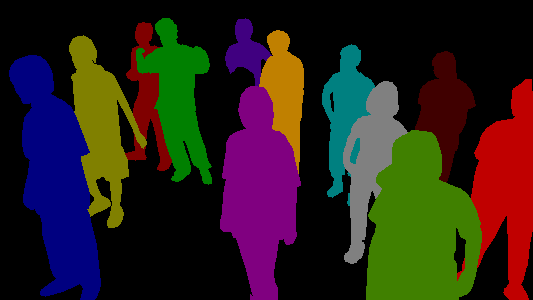}
  \includegraphics[width=0.25\linewidth]{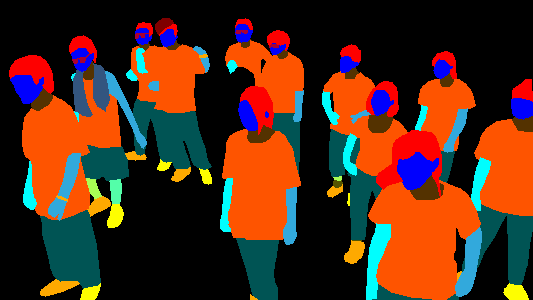} 
 \includegraphics[width=0.25\linewidth]{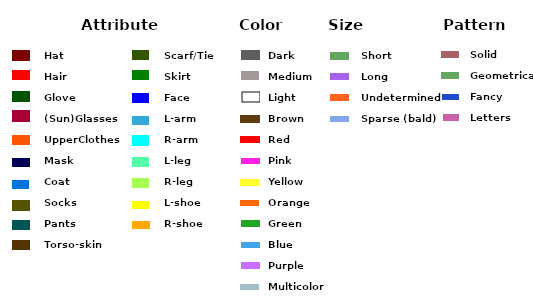}  
   \includegraphics[width=0.25\linewidth]{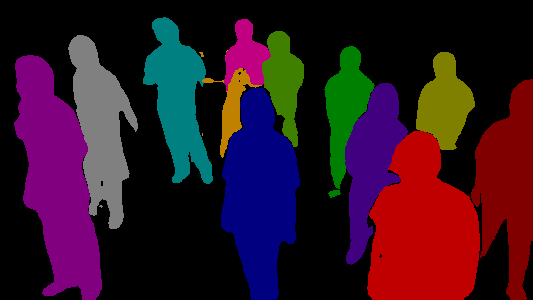}
   \includegraphics[width=0.25\linewidth]{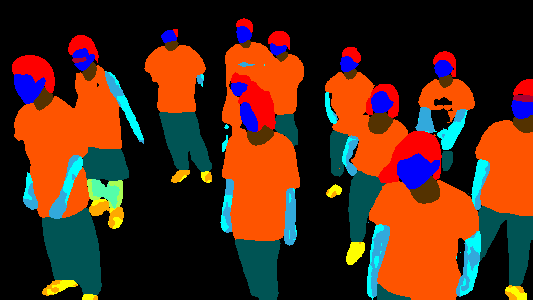} 
  \includegraphics[width=0.25\linewidth]{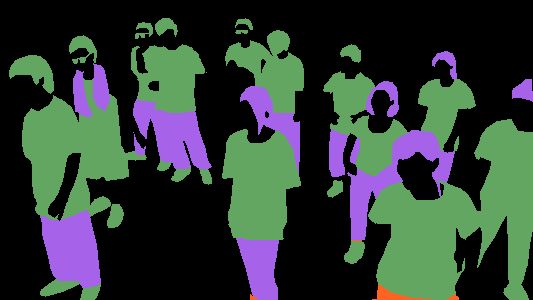}   \includegraphics[width=0.25\linewidth]{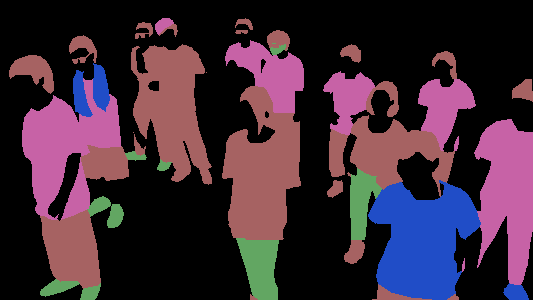} \includegraphics[width=0.25\linewidth]{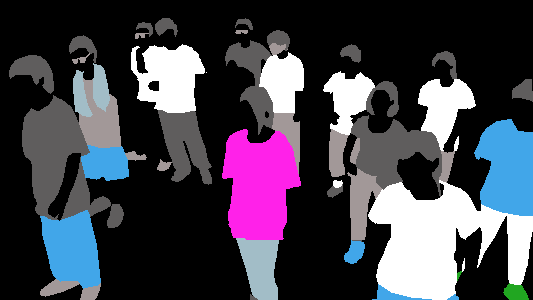} \includegraphics[width=0.25\linewidth]{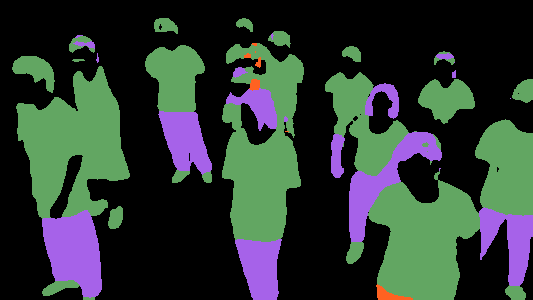} 
 \includegraphics[width=0.25\linewidth]{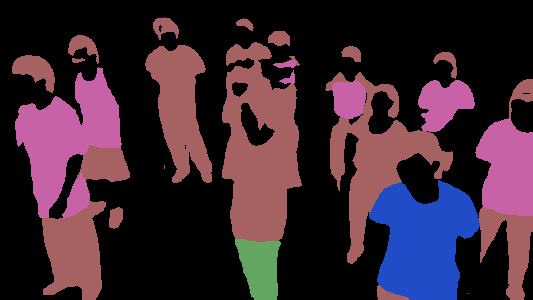}   
  \includegraphics[width=0.25\linewidth]{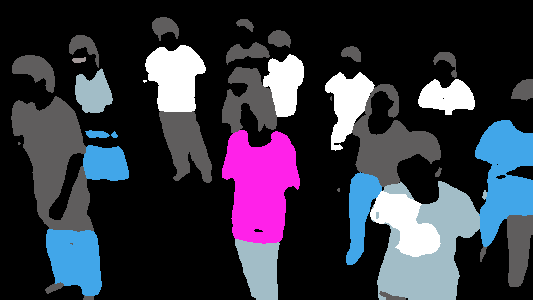}
  \caption{
  Example of \cihpc{} ground truths 
  (1st and 3rd rows) 
  and  our model predictions 
  (2nd and 4th rows).  
  First two rows: RGB image and legends, human instance, and semantic attribute maps. Last two rows: size, pattern and color maps.
  }
  \label{fig:prediction}
\end{figure}

Public datasets have been proposed for \hp{} (cf. details in~\cite{Cheng2020}).
Many of them target retail/fashion applications and, thus, gather single-person images in controlled environments~\cite{Dong2013, Yamaguchi2013, Yang2014, 
Liu2013, 
Zheng2018, 
Ge2019}. 
Others present multiple-person images taken under in-the-wild environments~\cite{Xia2017, Li2017b, Zhao2018, Gong2018}.
However, detecting the presence of attributes (e.g. pants, coat) is, in general, not sufficient to describe a person distinctively. 
Attributes should be characterized to provide a more complete and useful description. 
But no available dataset provide localized attributes with characteristics (such as color, size and pattern) in multi-person images.
Indeed, even if some fashion-aimed datasets~\cite{Yang2014, Liu2013, Zheng2018, Ge2019} provide some color and pattern tags, they target single-person images with good resolution and pose.
MHP v2.0~\cite{Zhao2018} provides many fine-grained attributes which partially include the size characteristic (e.g. pants vs. shorts, boot vs. shoe) but no color or pattern characteristics.
This encouraged us to create a new \hp{} dataset with multi-person characterized attributes. Thus, we have worked on CIHP~\cite{Gong2018}, the largest existing multi-\hp{} dataset, and 
annotated characteristics.

In order to cope with characterized \hp{}, we also propose a method to serve as a baseline. 
Among existing \hp{} methods,
\emph{single-person} methods make the assumption of a single person in the image, which means they do semantic segmentation and do not manage the attribute-to-person assignment problem~\cite{Gong2019, He2020}. 
In contrast, \emph{multi-person} methods cope with this problem by doing instance-level \hp{}. They can be divided into three main categories. 
\emph{Top-down two-stage} methods require human-instance segmentation of the image as an additional input~\cite{Li2017a,Ji2019,Liu2019,Ruan2019}. 
Their computation time highly depends on the number of people in the image. 
\emph{Top-down one-stage} methods ~\cite{Qin2019,Yang2019,
Yang2020b} predict both human instances and attributes.
But computation time still depends of the number of people in the image because the top-down strategy forces to forward in the local parsing branch(es) as many times as the number of ROI candidates (people). 
%
\emph{Bottom-up} methods also predict both human instances and attributes.
Yet, their computation time does not depend on the number of people in the image~\cite{Li2017b,Zhao2018,Gong2018}.
However, these methods generally rely on expensive test-time augmentations, post-processing~\cite{Li2017b,Gong2018} or heavy GAN architectures~\cite{Zhao2018}. 
\emph{Low and constant computation time} is essential when
applying multi-\hp{} on large amounts of possibly crowded images or videos.
Thus, we propose a fast end-to-end bottom-up model based on transformers, which 
manages also the 
new
task of attribute characterization, 
with low and constant computation time. 

The contributions of this article are two-fold: 
(1) a new dataset for multi-\hp{} in the wild with characteristics of localized attributes; 
(2) a bottom-up multi-task model for instance-level \hp{} with 
characterized attributes. The proposed model does not need post-processing.
It has low constant processing time, whatever the number of people per image, which makes it scalable and 
deployable.
We hope this new dataset and baseline will encourage research for fast and accurate methods for more complete human descriptions.

\vspace{-0.3cm}
\section{Proposed Dataset}
\label{sec:cihp++}
\begin{figure}[!h]
  \centering
  \includegraphics[width=0.8\linewidth]{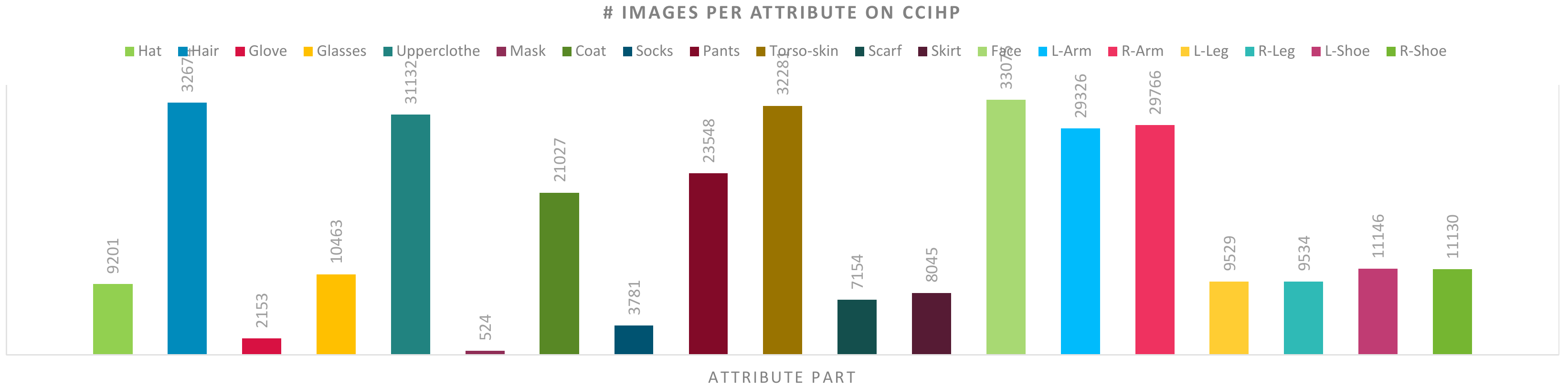}
  \includegraphics[width=0.8\linewidth]{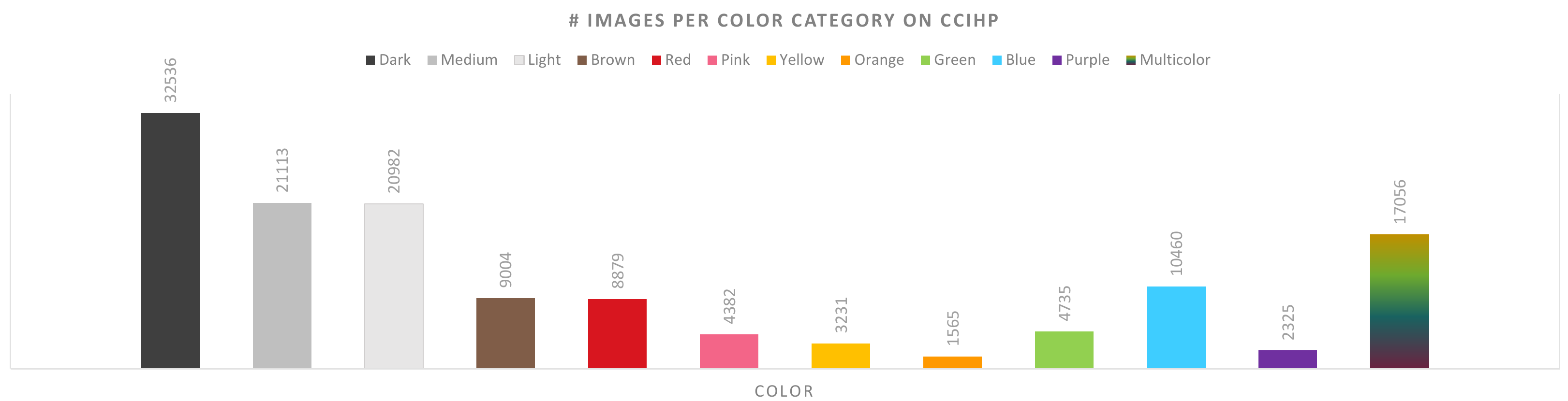}
  \includegraphics[width=0.8\linewidth]{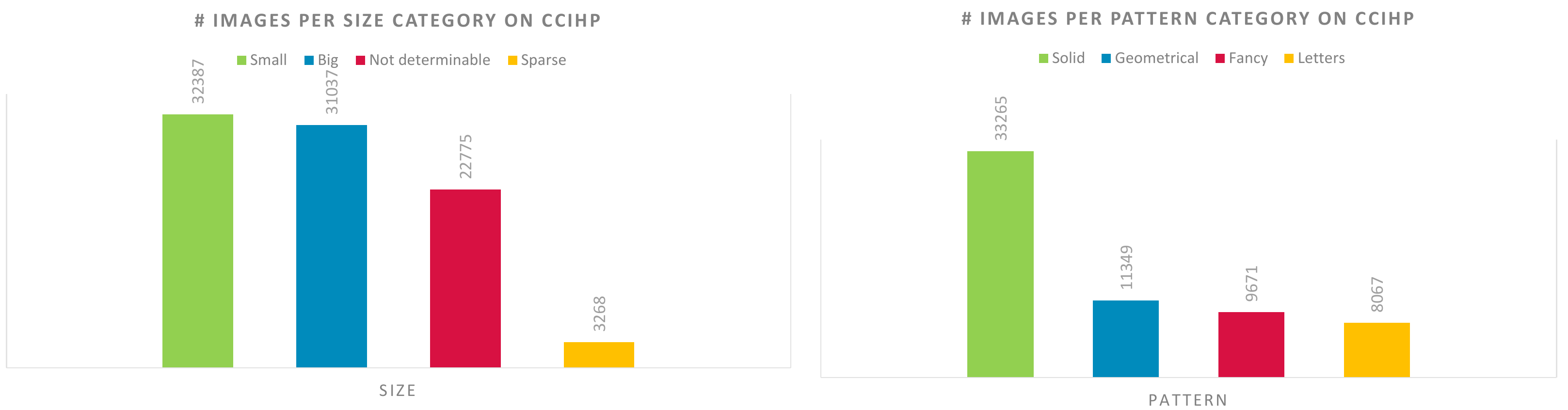}
  \caption{Image distribution on the 19 (resp. 12, 4, 4) semantic attribute (resp. color, size, pattern) labels in \cihpc{}.}
  \label{fig:stats}
\end{figure}

Our new dataset \cihpc{} (\emph{Characterized CIHP}) is based on the CIHP images~\cite{Gong2018}. 
We have kept 
the partition of the 33,280 images into 28,280 images for training and 5,000 for validation and testing.
However, small changes have been made on human instance masks and 
the 19
semantic attribute classes. 
Moreover, 
20
characteristic classes 
have been
annotated in a pixel-wise way. The first and third rows of Fig. \ref{fig:prediction} show an example of the new \cihpc{} annotations. More examples can be found on the supplemental material.
Fig.~\ref{fig:stats} shows the distribution of images per label.

\vspace{-0.5cm}
\paragraph*{Human instances}
110,821 people (+121 humans compared to CIHP \cite{Gong2018}) are annotated with full masks including accessories that 
did
not have labels in the semantic attributes. Thus, belt or necklace are for example integrated into the human masks. The number of people per image is still around 3.

\vspace{-0.5cm}
\paragraph*{Semantic attributes}
We have modified the CIHP classes but we still get 19 classes of clothing and body parts in the end. 
`Glasses' class now
includes sunglasses, glasses and eyewears.
`Scarf' class is extended to all clothing worn around the neck:
tie, bow-tie, scout scarf, ... 
Masks of the `Dress' class are split into the `UpperClothes' and `Skirt' classes.
Masks of `Hair' class are augmented with facial hair like beard and mustache. 
Finally, 
a new `Mask' class 
is added for all 
facial masks.

\vspace{-0.4cm}
\paragraph*{Size characteristics}
Four size classes 
characterize clothing attributes and/or hair: `Short/small', `Long/large', `Undetermined' (when the attribute in truncated or occluded by another one), `Sparse/bald' (for `Hair' class only). 
Combining attribute classes with size characteristic 
gives fine-grained attribute labels.
E.g., `Pants' + `Short' = shorts; `Shoes' + `Long' = boots.

\vspace{-0.4cm}
\paragraph*{Pattern characteristics}
Four pattern classes 
characterize clothing attributes and hair: `Solid', `Geometrical', `Fancy', `Letters'. 
The `Solid' class is 3 times more represented than the others (cf. Figure~\ref{fig:stats}) 
as all `Hair' labels and 
lots of
clothing are solid.

\vspace{-0.4cm}
\paragraph*{Color characteristics}
Twelve color classes 
characterize clothing attributes and hair: `Brown', `Red', `Pink', `Yellow', `Orange', `Green', `Blue', `Purple',
`Multicolor' (when several colors are evenly represented on the attribute part), and colors with no specific hue, `Dark' (black to gray), `Medium' (gray), `Light' (gray to white).
E.g.: `Glasses' + `Dark' = sunglasses; `Glasses' + `Medium' = reading glasses.

\vspace{-0.3cm}
\section{Proposed Method}
\label{sec:method}

With this new dataset, we propose an original baseline called \hptr{} for \emph{Human Parsing with TRansformers}. Our approach is bottom-up and multi-task, sharing features between the different tasks to be 
scalable.

\vspace{-0.3cm}
\paragraph*{\hptr{} Architecture}
\begin{figure}[!t]
  \centering
  \includegraphics[width=0.8\linewidth]{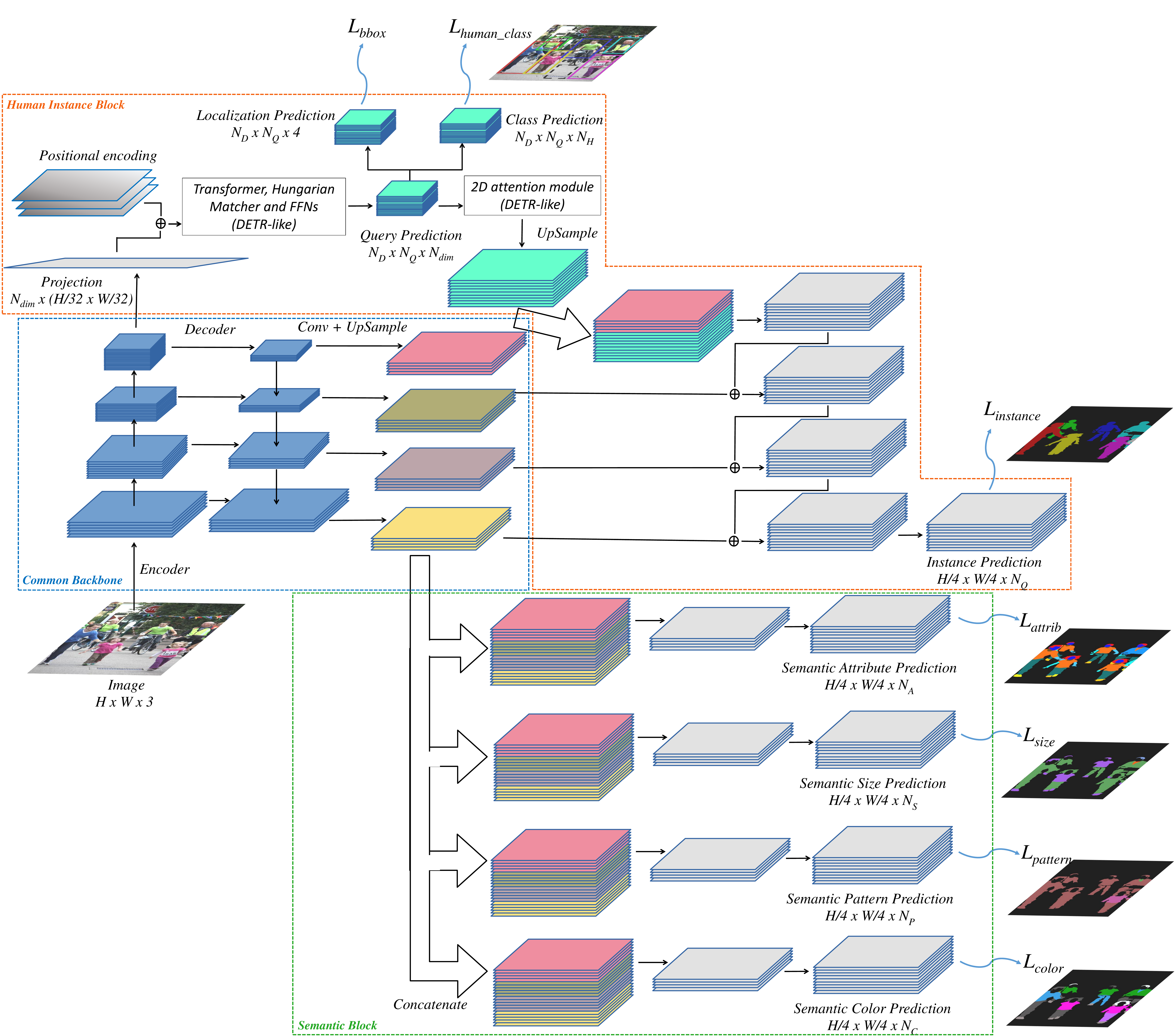}
  \caption{\hptr{} overview. From an RGB 
  $H$x$W$ image,
  our multi-task bottom-up model predicts semantic maps with $N_{A}$ (resp. $N_{S}$, $N_{P}$, $N_{C}$) channels for the $N_{A}$ (resp. $N_{S}$, $N_{P}$, $N_{C}$) learned attribute (resp. size, pattern, color) classes. With the use of transformers, human instance maps are also predicted with $N_{Q}$ channels. $N_{Q}$ 
  is also
  the number of queries 
  in
  the transformers's decoder. 
  $N_{dim}$, $N_{D}$ are hidden transformer dimensions set as in DETR \cite{Carion2020}.}
  \label{fig:overview}
\end{figure}
can be split into 3 blocks
(cf. Fig.~\ref{fig:overview}).
\\
(1) The \emph{Common Backbone} 
will share its features with all the task branches. It is composed of an encoder and a decoder 
to keep information at multiple resolutions,
like in detector architectures.
As in PandaNet 3D human pose estimation model
\cite{Benzine2020}, 
each pyramid feature map of the decoder is passed through 4 convolutional layers and resized to the size of the highest resolution level. These feature maps 
feed
the different task branches of the second and third blocks.
\\
(2) 
In the \emph{Human Instance Block}, 
we exploit transformers for human detection as in the recent DETR 
approach 
(see details in~\cite{Carion2020}),
to ensure that our architecture keeps bottom-up property and does not require post-processing at the end of the inference.
However, the mask branch of our \emph{Human Instance Block} differs from DETR: 
To share a maximum of features with the other branches of our model, we do not generate instance mask from feature maps of the encoder.
Instead, we
concatenate the up-scaled feature maps of the decoder with the feature maps from the transformer's outputs. 
It is achieved by creating a top-down pathway with lateral connections to bottom-up convolutional layers.
In the end, this block provides human bounding boxes, 
scores and masks.
\\
(3) \emph{Semantic Block} 
corresponds to the semantic attribute and characteristic (size, pattern, color) branches. 
This block allows a pixel-wise segmentation of people body parts, clothing and their characteristics.
\emph{Semantic Block} 
is made of 
4 similar branches, one per task. 
Each of them 
is fed 
with the concatenation of the up-scaled pyramid feature maps of the decoder. These maps are forwarded into a convolutional semantic head to predict the semantic attribute, size, pattern and color outputs
according to the classes defined in \cihpc{}.

\vspace{-0.3cm}
\paragraph*{Training objective}

To train the \emph{Human Instance Block}, we follow DETR \cite{Carion2020} and use the same objective. The training loss $\mathcal{L}_{human}$ is the sum of $\mathcal{L}_{Hungarian}$ (not presented in Fig. \ref{fig:overview}), $\mathcal{L}_{bbox}$ (an $l_1$ loss), $\mathcal{L}_{human\_class}$ (a cross-entropy loss), and $\mathcal{L}_{instance}$ (composed of a DICE loss \cite{Milletari2016} and a Focal loss \cite{Lin2017}). 
In the \emph{Semantic Block}, we also use for each head a DICE loss and a Focal loss 
(rather than a cross-entropy loss)
to better deal with imbalanced classes. 
No weighting is used between each loss. The global training objective is then as follows: 
$\mathcal{L} = \mathcal{L}_{human} + \mathcal{L}_{attrib}$ + $\mathcal{L}_{size}$ + $\mathcal{L}_{pattern}$ + $\mathcal{L}_{color}$.

\vspace{-0.1cm}
\section{Experiments and Results}
\label{sec:Experiments}
%
%
%

\paragraph*{Implementation Details}
%
\hptr{} has been implemented in Pytorch. 
The encoder 
is a ResNet50 
pre-trained on ImageNet dataset. The \emph{Human Instance} and \emph{Semantic Blocks} are trained jointly for 300 epochs on 8 Titan X (Pascal) GPUs with 1 image per batch.
The long side of training and validation images can not exceed 512 pixels. 
$N_Q$ is set to 40 person queries per image. 
Please refer to \cite{Carion2020} for other hyper-parameter settings and input transformations.

\begin{figure}[!b]
  \centering
  \includegraphics[width=0.55\linewidth]{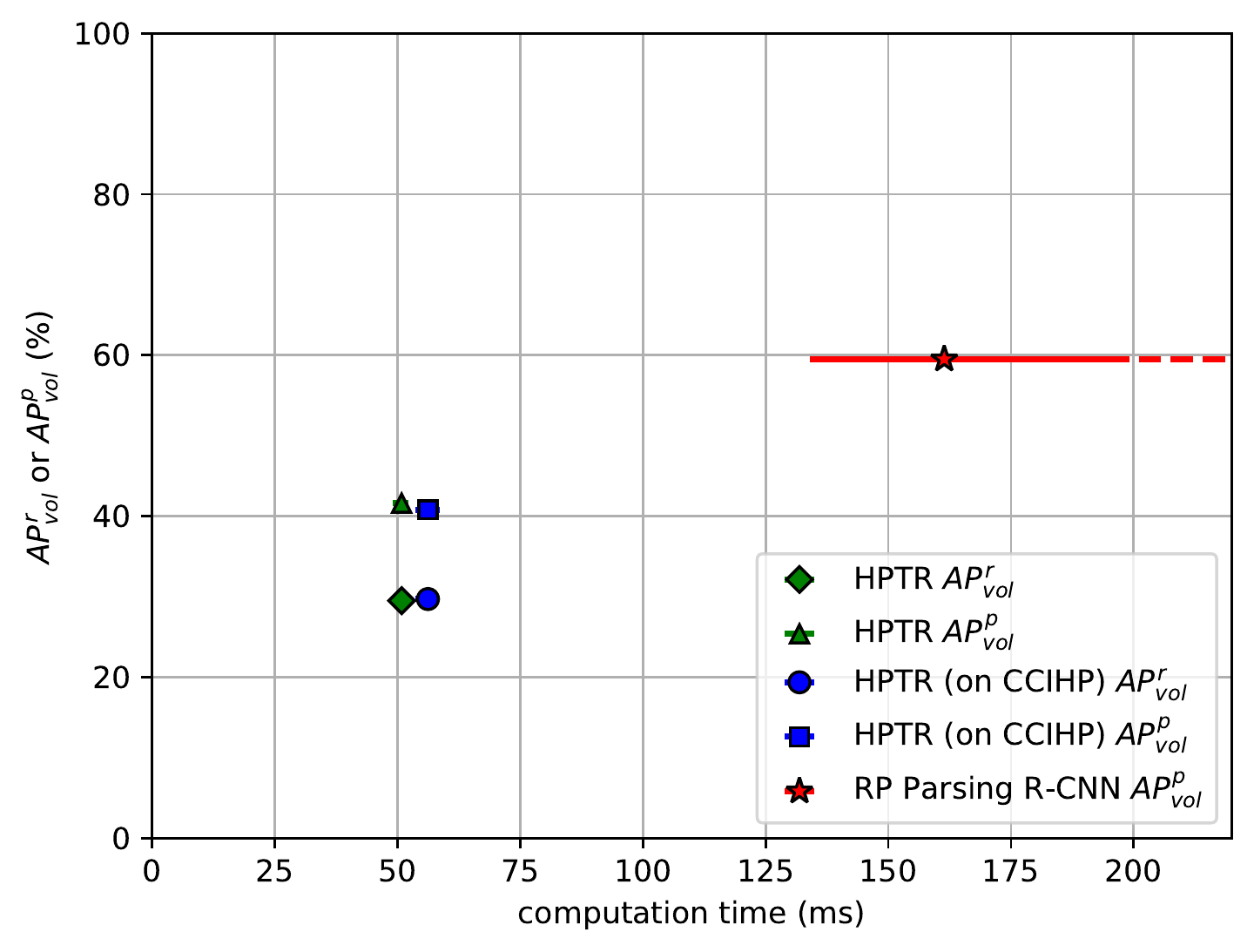}
  \caption{
  Precision/time trade-off of \hptr{} on CIHP and \cihpc{} (on a TITAN X GPU), compared to RP Parsing R-CNN that achieved best precision and speed in multi-\hp{} state of the art. \hptr{} is the fastest method and has constant time while RP Parsing R-CNN (a top-down approach) is more accurate but slower and not scalable: red horizontal solid line shows time variation with 2 to 18 people per image.
 %
}
  \label{fig:times}
\end{figure}

%
\begin{table*}[!t]
\centering
\begin{minipage}[!t]{0.99\linewidth}
\subfloat[][\label{tab:attrib_per_class}]{
\resizebox{\linewidth}{!}{
\begin{tabular}{|l|*{22}{c}*{1}{c|}}
 \cline{2-24}
\multicolumn{1}{c|}{} & \multicolumn{1}{c|}{\small{all}} & \small{Hat}& \small{Hair} &  \small{Glove} & \small{Sunglasses} & \small{Glasses} & \small{UpperClothes} & \small{Dress} & \small{Mask} & \small{Coat} & \small{Socks} & \small{Pants} & \small{Torso-skin} & \small{Scarf} & \small{Scarf/Tie}& \small{Skirt} & \small{Face} & \small{L-arm} & \small{R-arm} & \small{L-leg} & \small{R-Leg} & \small{L-shoe} & \small{R-shoe} \\\hline
\small{CIHP mIoU} & \multicolumn{1}{c|}{\small{53.0}} & \small{58.1}& \small{70.4} &  \small{10.5} & \small{33.9} & \small{-} & \small{53.5} & \small{34.6} & \small{-} & \small{41.2} & \small{17.3} & \small{49.2} & \small{57.2} & \small{7.6}& \small{-} & \small{19.5} & \small{69.0} & \small{38.9} & \small{31.7} & \small{24.1} & \small{18.8} & \small{20.1} & \small{16.9} \\\hhline{========================}
\small{\cihpc{} mIoU} & \multicolumn{1}{c|}{\small{52.1}} & \small{55.0}& \small{69.7} &  \small{7.5} & \small{-} & \small{41.0} & \small{48.9} & \small{-} & \small{8.2} & \small{44.7} & \small{16.8} & \small{50.0} & \small{57.8} & \small{-} & \small{34.1} & \small{38.0} & \small{66.0} & \small{38.8} & \small{30.2} & \small{23.7} & \small{19.5} & \small{19.5} & \small{17.1} \\\hline
\small{\cihpc{} \aprv{}} & \multicolumn{1}{c|}{\small{29.7}} & \small{52.0}& \small{60.9} &  \small{6.6} & \small{-} & \small{32.1} & \small{45.1} & \small{-} & \small{10.7} & \small{48.3} & \small{12.8} & \small{52.8} & \small{48.1} & \small{-} & \small{28.0} & \small{40.3} & \small{71.3} & \small{9.0} & \small{11.6} & \small{10.8} & \small{8.4} & \small{7.2} & \small{7.2} \\\hline
\end{tabular}
}
}
\end{minipage}
\begin{minipage}[!t]{0.99\linewidth}
\subfloat[][\label{tab:charac_per_class}]{
\resizebox{\linewidth}{!}{
\begin{tabular}{|l|*{12}{c}*{1}{c||}*{4}{c}*{1}{c||}*{5}{c}}
 \cline{2-24}
\multicolumn{1}{c|}{} & \multicolumn{13}{c||}{semantic color} & \multicolumn{5}{c||}{semantic size} & \multicolumn{5}{c|}{semantic pattern}\\ \cline{2-24}
\multicolumn{1}{c|}{} & \multicolumn{1}{c|}{\small{all}} & \small{Dark} & \small{Medium} & \small{Light} & \small{Brown} & \small{Red} & \small{Pink} & \small{Yellow} & \small{Orange} & \small{Green} & \small{Blue} & \small{Purple} & \small{Multicolor} & \multicolumn{1}{c|}{\small{all}} & \small{Short} & \small{Long} & \small{Undet.} & \small{Sparse} & \multicolumn{1}{c|}{\small{all}} & \small{Solid} & \small{Geom.} & \small{Fancy} & \multicolumn{1}{c|}{\small{Letters}} \\\hline
\small{mIoU} & \multicolumn{1}{c|}{\small{43.8}} & \small{60.3} & \small{23.1} & \small{31.5} & \small{9.0} & \small{19.7} & \small{11.4} & \small{11.2} & \small{1.1} & \small{12.4} & \small{15.7} & \small{4.9} & \small{14.2} & \multicolumn{1}{c|}{\small{58.8}} & \small{55.0} & \small{58.6} & \small{20.5} & \small{14.6} & \multicolumn{1}{c|}{\small{67.4}} & \small{72.3} & \small{30.0} & \small{21.0} & \multicolumn{1}{c|}{\small{16.3}} \\\hline
\small{\apcrv{}} & \multicolumn{1}{c|}{\small{15.0}} & \small{40.8} & \small{18.4} & \small{25.2} & \small{5.5} & \small{17.4} & \small{9.1} & \small{9.6} & \small{0.07} & \small{14.8} & \small{22.3} & \small{3.2} & \small{13.2} & \multicolumn{1}{c|}{\small{24.5}} & \small{33.1} & \small{37.5} & \small{13.5} & \small{13.7} & \multicolumn{1}{c|}{\small{20.9}} & \small{36.9} & \small{14.4} & \small{14.1} & \multicolumn{1}{c|}{\small{18.2}} \\\hline
\end{tabular}
}
}
\end{minipage}
\caption{\hptr{} performances per class in \%. (a) Results on CIHP and \cihpc{} attributes. (b) Results on \cihpc{} color, size and pattern characteristics (Undet. stands for undetermined and Geom. for Geometrical).}
\label{fig:miou_per_class}
\end{table*}
\vspace{-0.3cm}
\paragraph*{Datasets and Evaluation Protocols}
\hptr{} is evaluated on proposed dataset \cihpc{}, as a first baseline of characterized multi-\hp{}, and also on CIHP~\cite{Gong2018} for comparison with the state of the art of multi-\hp{} without characterization.
%
%
%
We follow PGN \cite{Gong2018}  evaluation implementation by using \emph{mean Intersection over Union (mIoU)} \cite{Long2015} for attribute and characteristic (size, pattern, color) semantic segmentation evaluation. As instance-level \hp{} evaluation metric, we use \emph{mean Average Precision based on region (\aprv{})} \cite{Hariharan2014}. 
We also compute \emph{mean Average Precision based on part (\appv{})} \cite{Zhao2018} following Parsing R-CNN \cite{Yang2019} evaluation protocol. 
Finally, we extend \aprv{} metric 
to evaluate attribute characterization: 
Whereas \aprv{} evaluates the prediction of attribute (class \& score) relative to each instanced attribute mask, \emph{mean Average Precision based on characterized region (\apcrv{})} evaluates the prediction of characteristic (class \& score) relative to each instanced \emph{and} characterized attribute mask, \emph{independently} of the attribute class prediction. 
Thus, \apcrv{} is jointly conditional on human instance segmentation, attribute delineation and characteristic segmentation.


\vspace{-0.3cm}
\paragraph*{Comparison with state-of-the-art of  multi-\hp{} without characterization}
\label{ssec:hp_evaluation}

As multi-\hp{} methods do not manage attribute characterization proposed in \cihpc{}, we first evaluate \hptr{} without characterization task, on CIHP dataset~\cite{Gong2018}, to give an idea of its speed vs precision trade-off relative to these methods.
We compare \hptr{} with best methods of each approach family that have available models.
Inference time is averaged after 50 runs 
on a Titan X GPU, using 
CIHP images containing from 2 to 18 people.
\emph{Top-down} approaches reach the state-of-the art precision thanks to their local parsing branches:
\emph{two-stage} M-CE2P~\cite{Ruan2019} reports \aprv{} = 42.8\% and \emph{one-stage} RP Parsing R-CNN~\cite{Yang2020b} gets \appv{} = 59.5\%.
However, top-down approaches are known to have computation times dependent on the number of people in a scene and not be easily scalable. Typically, on CIHP images, M-CE2P (resp. RP Parsing R-CNN) runs in 752~ms--6.6~s (resp. 136--195~ms) according to the number (2--18) of people per image. By extrapolation to 40 people, this time would go beyond 14~s for M-CE2P and 285~ms for RP Parsing R-CNN.
In contrast, \emph{bottom-up} approaches, 
generally less accurate,
have the advantage to
run in constant time, independent of
the number of people. 
NAN~\cite{Zhao2018} runs in about 275~ms (but no $AP$ on CIHP was reported). As for PGN~\cite{Gong2018}, it reaches $AP^r_{vol}$ = 33.6\%,  and \appv{} = 39.0\%, in around 1.4~s without counting additional post-processing time.
These models are more than 5 and 29 times slower than HPTR. Indeed, our bottom-up approach 
has a good speed/precision trade-off with a low constant time of around 50 ms and precision similar to PGN 
(\aprv{} = 29.5\%, \appv{} = 41.6\%).
Fig.~\ref{fig:times} shows the speed/precision trade-off for the most precise method (RP Parsing RCNN) and the proposed \hptr{} which is the fastest method of the state of the art while having precision comparable to the most precise bottom-up method (PGN).
Low and constant computation time is essential 
when applying multi-\hp{} 
on large amounts of possibly crowded videos.

\vspace{-0.4cm}
\paragraph*{Results of multi-\hp{} with characterization}
Now, \hptr{} is evaluated on \cihpc{} and gets an overall mIoU of 52.1\%, \aprv{} of 29.7\% and \appv{} of 40.8\% (cf. Tab.~\ref{tab:attrib_per_class} for results detailed per attribute class).
%
mIoU results are also presented for CIHP with small differences in classes. Per class results are close to those obtained on CIHP. The main differences are on `Glasses', `Scarf/Tie' and `Skirt' classes that are more represented in \cihpc{} (cf. Sec. \ref{sec:cihp++}).
Tab. \ref{tab:charac_per_class} shows the mIoU and our new metric \apcrv{} for all characteristic classes. 
Overall, \hptr{} reaches an mIoU of 43.8\% (resp. 58.8\% and 67.4\%) and 
an \apcrv{} of 15.0\% (resp. 24.5\% and 20.9\%) 
for the color (resp. size and pattern) labels.
We can see 
that results are directly correlated to 
class frequency
(cf. Fig.~\ref{fig:stats}).
So we 
would like to address
this class imbalance issue as future work to improve \hptr{} performance.
Consistency 
between 
characteristic 
and attribute masks, 
observed qualitatively as in Fig.~\ref{fig:prediction},
is driven by
the backbone features shared by the task branches.
Besides, computation time of 56 ms shows that \hptr{} is also scalable with these 3 additional characterization tasks (compared to 50 ms without characterization, cf. Fig. \ref{fig:times}).
Thus, the use of global branches for additional tasks, instead of local branches, gives 
another advantage over top-down approaches.

\vspace{-0.3cm}
\section{Conclusion}
\label{sec:conclusion}
In this article, we propose \cihpc{}, the first multi-\hp{} dataset with systematic characterization of instance-level attributes.
It is based on CIHP images, the largest existing multi-\hp{} in-the-wild dataset.
We have defined 20 classes of characteristics split into 3 categories (size, pattern and color) to better describe each human attribute.
To learn these characteristics, we have developed \hptr{}, a bottom-up, 
and multi-task baseline. It has low and constant computation time.
Thus, it is scalable with the number of people per image and the number of tasks to learn.
We hope that research towards fast and accurate methods for more complete human descriptions  will be encouraged thanks to this  new  dataset  and baseline.

\vspace{-0.4cm}
\paragraph*{Acknowledgments}
This publication was made possible by the use of the FactoryIA supercomputer, financially supported by the Ile-de-France Regional Council.


\bibliographystyle{IEEEbib}
\bibliography{version_1}

\end{document}